\documentclass{article}

\usepackage{PRIMEarxiv}

\usepackage[utf8]{inputenc} % allow utf-8 input
\usepackage[T1]{fontenc}    % use 8-bit T1 fonts
\usepackage{hyperref}       % hyperlinks
\usepackage{url}            % simple URL typesetting
\usepackage{booktabs}       % professional-quality tables
\usepackage{amsfonts}       % blackboard math symbols
\usepackage{nicefrac}       % compact symbols for 1/2, etc.
\usepackage{microtype}      % microtypography
\usepackage{lipsum}
\usepackage{fancyhdr}       % header
\usepackage{graphicx}       % graphics
\graphicspath{{media/}}     % organize your images and other figures under media/ folder

\title{Quality-Constant Per-Shot Encoding by Two-Pass Learning-based Rate Factor Prediction
}

\author{
  Chunlei Cai, Yi Wang, Xiaobo Li, Tianxiao Ye \\
  Bilibili Inc. \\
  Shanghai, China \\
  \texttt{\{caichunlei, wangyi, lixiaobo, yetianxiao\}@bilibili.com} \\
}

\begin{document}
\maketitle

\begin{abstract}
Providing quality-constant streams can simultaneously guarantee user experience and prevent wasting bit-rate. 
In this paper, we propose a novel deep learning based two-pass encoder parameter prediction framework to decide rate factor (RF), with which encoder can output streams with constant quality. For each one-shot segment in a video, the proposed method firstly extracts spatial, temporal and pre-coding features by an ultra fast pre-process. Based on these features, a RF parameter is predicted by a deep neural network. Video encoder uses the RF to compress segment as the first encoding pass. Then VMAF quality of the first pass encoding is measured. If the quality doesn't meet target, a second pass RF prediction and encoding will be performed. With the help of first pass predicted RF and corresponding actual quality as feedback, the second pass prediction will be highly accurate. Experiments show the proposed method requires only 1.55 times encoding complexity on average, meanwhile the accuracy, that the compressed video's actual VMAF is within $\pm1$ around the target VMAF, reaches 98.88\%.
\end{abstract}

% keywords can be removed

\section{Introduction}
\begin{figure}[tbhp]
\setlength{\abovecaptionskip}{0.cm}
\setlength{\belowcaptionskip}{0.cm}
\centerline{\includegraphics[width=0.9\linewidth]{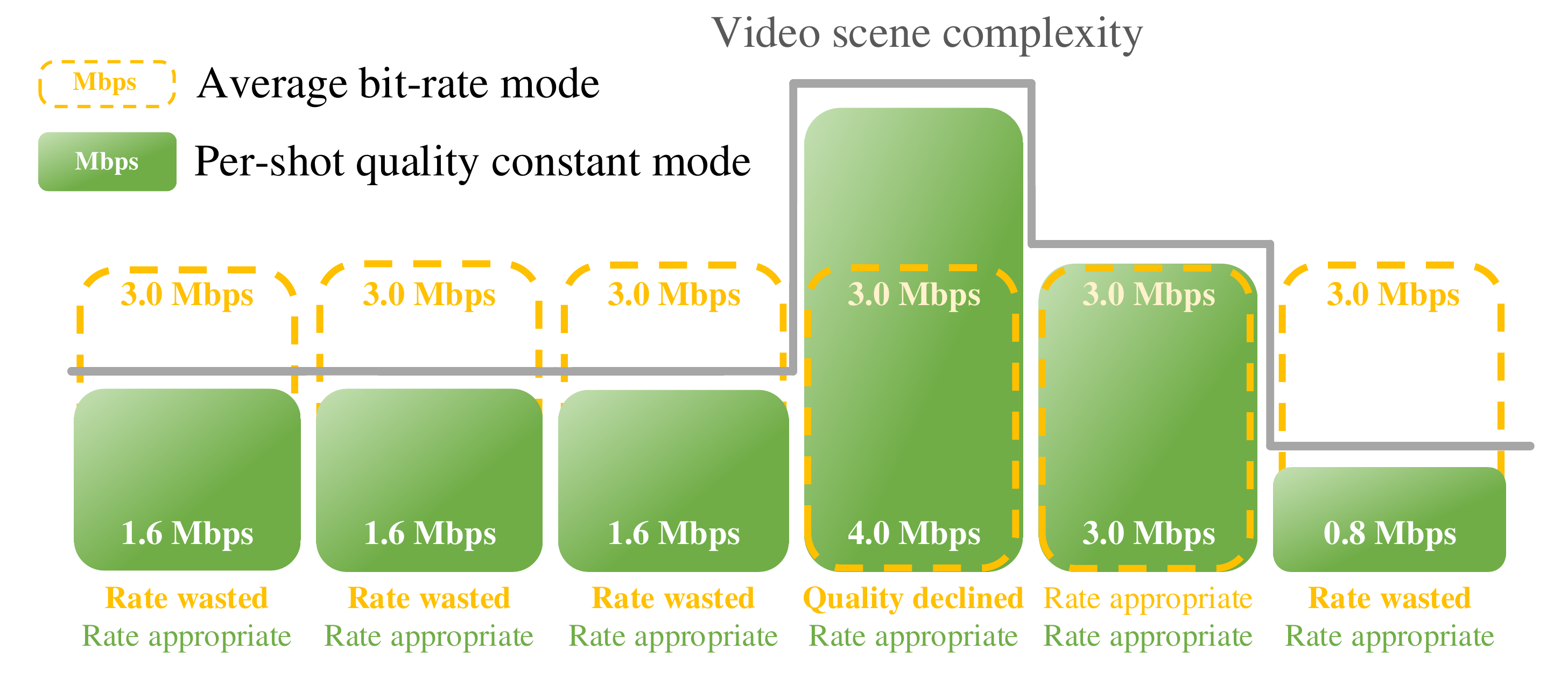}}
\caption{Per-shot quality constant coding mode can simultaneously guarantee quality and prevent wasting bit-rate.}
\label{fig:intro}
\end{figure}

Compared with other methods, such as average bit-rate mode, quality-constant compression can simultaneously guarantee quality and prevent wasting bit-rate as illustrated in Fig. \ref{fig:intro}.

% However, realizing a quality-constant compression method is not as simple as realizing an average bit-rate method. 
Realizing quality-constant compression requires addressing two difficulties. The first is how to measure compressed video quality that matches human subjective experience. The second is how to decide appropriate coding parameters with which an encoder can output compressed video with target quality. 

To address the first video quality measurement problem, Netflix proposed a metric named Video Multi-method Assessment Fusion (VMAF) \cite{li2018vmaf}, that has been widely used in the industry. Because VMAF shows high correlation with human subjective experience, this paper uses VMAF as video quality metric.

To address the second coding parameter decision problem, many methods have been proposed \cite{katsenou2021efficient,covell2016optimizing,xing2019predicting,sun2018machine}. For example, Covell et al. \cite{covell2016optimizing} proposed a neural network based method to predict RF parameter, with which encoder can compress video at required bit-rate under Constant Rate Factor (CRF) coding mode. This method achieves up to 80\% accuracy if a pre-coding is performed. Xing et al. \cite{xing2019predicting} proposed a deep learning based RF prediction method to compress video at required VMAF quality, which prediction accuracy is 77.6\%.

For quality-constant compression methods, the accuracy of coding parameter prediction is most important. Because the more accurate the prediction is, the more stable quality will be provided across all videos and the less bit-rate will be wasted. 

In order to further improve the parameter prediction accuracy for quality-constant compression. This paper proposes a novel deep learning based two-pass prediction framework to decide Rate Factor (RF), with which an encoder can output streams with required quality at high accuracy. 

Overall, the proposed method follows per-shot video coding framework \cite{manohara2018optimized}, which individually compresses each one-shot segment with different coding parameters. For each one-shot video segment, the proposed method will decide an RF corresponding to the required VMAF quality. So the whole compressed video will have constant quality between segments. 

\begin{figure}[tbp]
\setlength{\abovecaptionskip}{0.cm}
\setlength{\belowcaptionskip}{0.cm}
\centerline{\includegraphics[width=0.9\linewidth]{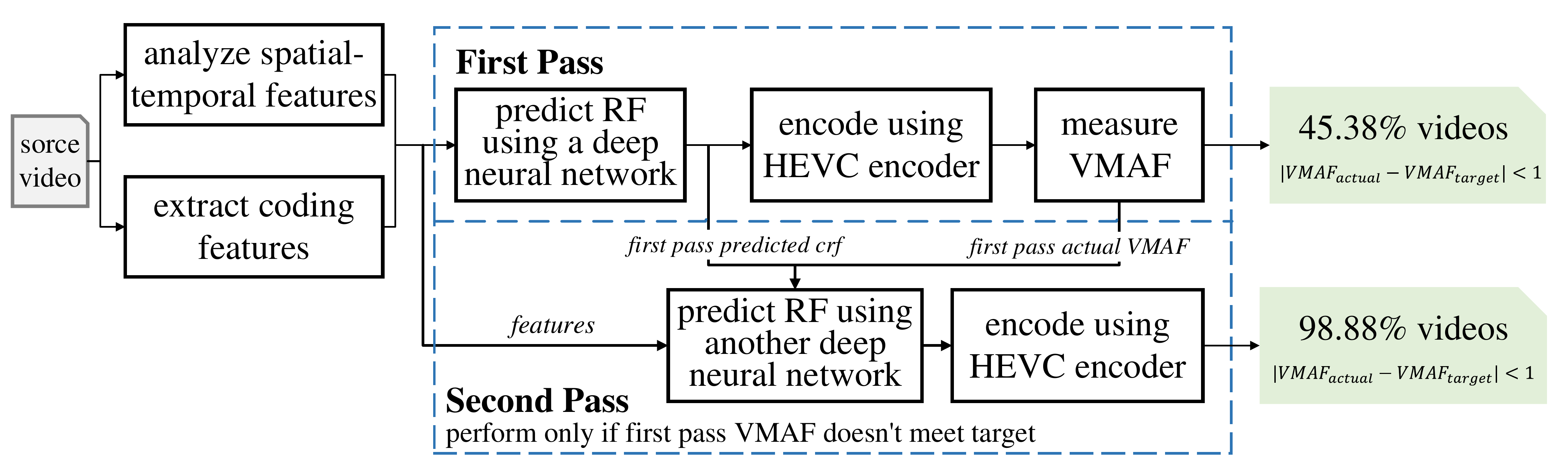}}
\caption{The proposed deep learning based two pass RF prediction framework, which reaches 98.88\% accuracy to compress video at target quality at only 1.55 times of encoding complexity on average.}
\label{fig:framework}
\end{figure}

In detail, the proposed method firstly splits a video into one-shot video segments by scene-cut algorithm used in x264 encoder \cite{x264}. For each one-shot segment, spatial-temporal and pre-coding features are extracted by an ultra fast pre-process. Based on these features, an RF parameter is predicted by a deep neural network. With the predicted RF, video encoder (such as HEVC encoder \cite{sullivan2012overview}) compresses the segment under CRF mode. This is the first encoding pass. Then VMAF quality of the first pass encoding is measured. If the VMAF value meets the target, then we store the compressed stream and move on to compress the next segment. 

Otherwise, a second pass RF prediction will be performed. This pass uses another deep neural RF prediction network. The input features of the second network composes of not only all features in the first pass, but also two other features: the first pass predicted RF and its corresponding actual VMAF value. After predicting the new RF, a second pass encoding is used to compress the segment. With the help of first pass results as feedback, the second pass prediction will be highly accurate, so we directly finish the current segment and move on to next one. Finally, the whole video is compressed, which all segments have same quality.

Experiments show the proposed method requires only 1.55 passes encoding and 1 pass VMAF measurement complexity on average, meanwhile the accuracy, that the percentage of the compressed video's VMAF within the target VMAF$\pm1$, reaches 98.88\%.

In the proposed method, the scene-cut algorithm of x264 encoder \cite{x264} is used to split a video into one-shot segments, so we mainly discuss how to predict RF for each segment in this paper. In the followings, we first introduce how to prepare video features for predicting RF, then present RF prediction deep neural network. Finally, we present experiment details.

\section{Video Features for Predicting RF}

In this paper, we extract spatial-temporal and pre-coding video features for predicting RF in the first pass. If the first-pass prediction and encoding don't generate stream with required quality, then the results of the first-pass are added to features, with which a second-pass prediction and encoding is further performed.

\subsection{Spatial and Temporal Features} \label{SubSec-Spatial-Tempotal-Features}

To represent video content characteristics, many spatial-temporal features have been proposed \cite{mohanaiah2013image,briechle2001template}. We apply Gray-level Co-occurrence Matrix (GLCM) and Normalized Correlation Coefficient (NCC), which are commonly used in coding parameter prediction methods \cite{katsenou2021efficient, katsenou2016predicting}.

GLCM can be used to represent spatial characteristics of video content \cite{mohanaiah2013image}. In detail, for each frame in a video segment, we compute 3 GLCM matrices, which distance of co-occurrence pixels are set as 1, 3 and 5 respectively. For each GLCM matrix of each frame, then we calculate 5 features, including energy, entropy, homogeneity, correlation and contrast. Then for each type of feature, we gather all features of all frames, and calculate mean and standard variance to represent the spatial characteristics of a whole video segment. Totally, a video segment has 30 ($3 \times 5 \times 2$) GLCM features.

NCC can be used to represent temporal characteristics of video \cite{briechle2001template}. For each pair of neighbour frames in a video segment, we compute a NCC matrix. For coefficients in each NCC matrix, we calculate 5 types of statistical values, including mean, standard variance, entropy, skew and kurtosis. For each type of statistical values of all pairs of frames, we compute mean and standard variance to represent temporal characteristics of a video segment. Totally, a video segment has 10 ($5 \times 2$) NCC features.

\subsection{Ultra-fast Pre-coding Features} \label{SubSec-Precoding-Features}

Besides spatial-temporal features, we also use a set of pre-coding features to represent video coding characteristics. To avoid introducing high computational complexity, we use Nvidia hard-ware H.264 encoder \cite{patait2016high} to pre-code video at an ultra-fast speed. Then we extract coding features from the pre-coded stream. 

Rather than pre-coding original video $Video$ directly, we reorganize the frames of original video and generate a new sequence of frames. For original video with total $n$ frames, we name its frames as $\{f_1, f_2, f_3, ... , f_{n_1}, f_n\}$. We firstly duplicate $\{f_2, f_3, ..., f_{n-2}, f_{n-1}\}$ and then reorganize all these frames into a new sequence of frames as $\{f_1, f_2, f_2, f_3, f_3, f_4, ..., f_{n-2}, f_{n-1}, f_{n-1}, f_{n}\}$. This new reorganized video ($Video^{'}$) has $2n-2$ frames. 

Then we encode $Video^{'}$ using a fixed Group Of Picture (GOP) setting as $\{IPIP...IPIP\}$. Under this setting, the frames are encoded as $\{[I(f_1), P(f_2)], [(I(f_2), P(f_3)],$ $...$ $, [I(f_{n-2})), P(f_{n-1})], [I(f_{n-1}), P(f_{n})]\}$. In another word, despite the first and last frames, all frames are encoded twice, one using Intra mode and another using Inter mode. When a frame is encoded using Intra mode, the coding features can represent spatial complexity. When a frame is encoded using Inter mode, the coding features can represent temporal complexity or correlation with previous frame. 

Afterwards, we extract and analyze coding features of each frame by parsing coded stream. For Intra coded frame we extract 34 features, including consumed bytes, ratio of each macro-block size ($4\times4$, $8\times8$ or  $16\times16$), ratio of each type of Intra prediction mode, etc. For Inter coded frame we extract 26 features, including consumed bytes, ratio of Intra/Inter modes, ratio of transform skip blocks, averaged length of motion vectors, etc. 

For each type of Intra coding features, we compute 5 types of statistical values between all Intra frames, including mean, standard variance, skew, kurtosis and entropy. Similarly, for each type of Inter coding features, we also compute 5 types of statistical values between all Inter frames. 

Finally, by this ultra-fast pre-coding process, we obtain total 300 ($34\times5 + 26\times5$) features to represent video coding characteristics.

\subsection{Feedback Features from First-pass}

After aforementioned video features are obtained, an RF is firstly predicted by a pre-trained deep neural network. With the predicted RF, we encode video and then measure the VMAF of the compressed video. If the actual VMAF doesn't meet target, a second-pass prediction will be performed. In the second-pass prediction, besides all the features used in the first-pass prediction, we add two extra features from the results of the first-pass. One is the predicted RF of first prediction and another is the actual VMAF of first encoding. 

With the help of the first pass results as anchor, the second pass will adjust the predicted RF automatically. If the first pass VMAF is higher than target, the second pass predicted RF will get larger. In contrast, if the first pass VMAF is lower than target, the second predicted RF will get smaller. Based on another pre-trained deep neural network, the second pass prediction will achieve high accuracy up to 98.88$\%$, which will be shown in Experiments Chapter.

\section{RF Prediction Deep Neural Network}

Based on aforementioned several hundreds of features, we need to solve a prediction problem, in where an appropriate RF, using which encoder can compress video with required VMAF quality, should be predicted accurately. To solve this problem, this paper build deep neural networks and learn parameters from large scale of data. The proposed deep neural network is shown as Fig. \ref{fig:intro}. 

\subsection{Network Architecture}

This network composes of several essential modules, such as batch normalization \cite{ioffe2015batch}, attention blocks \cite{vaswani2017attention} and res-net blocks \cite{dong2014learning}. Each module plays important role.

Elements from the input features have different properties and various distributions. For example, among coding features, the counts of bytes consumed for each frame are positive integers, while the ratio of each prediction mode is continuous real number ranged in $[0, 1]$. The difference of property and distribution between features makes training neural networks hard to converge \cite{ioffe2015batch}. To address this obstacle, this paper uses batch normalization, which was proposed to eliminate property and distribution differences between features and to  improve training performance.

\begin{figure}[tbp]
\setlength{\abovecaptionskip}{0.cm}
\setlength{\belowcaptionskip}{0.cm}
\centerline{\includegraphics[width=0.9\linewidth]{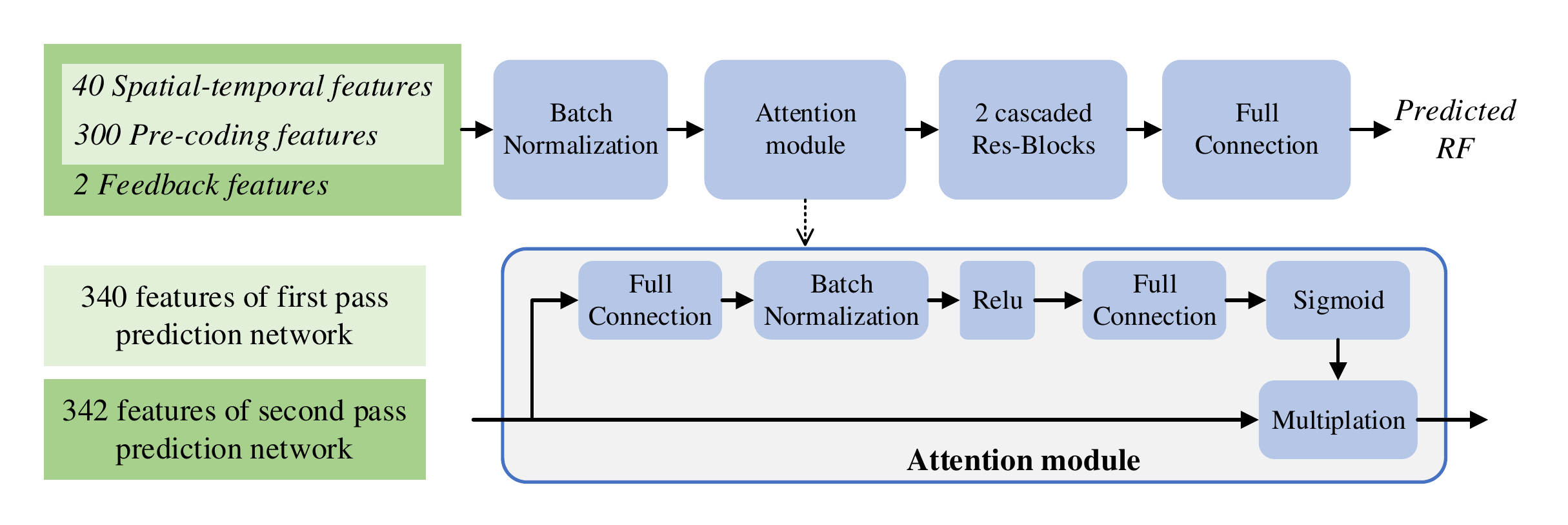}}
\caption{The proposed deep neural RF prediction networks. Note that first and second pass prediction networks are two individual models, which although have similar backbone architecture but different input features and individual parameters.}
\label{fig:network}
\end{figure}

Among all the input features, some contribute to prediction accuracy a lot but some others may contribute much less. Those useless features improve the training difficulties, decrease prediction accuracy and reduce generalization. To automatically select discriminative features, this paper applies Attention module \cite{vaswani2017attention}. In Attention module, a weight mask will be learned and multiplied to features. Those important features will be enlarged and affect result more, while those useless features will be suppressed and affect result less. This process will be learned automatically within the whole optimization of prediction accuracy. 

After normalization and Attention modules, several res-net blocks \cite{dong2014learning} are inserted. Res-net was proposed to improve training efficiency of deep networks and it has been widely used. So we apply this module to improve model performance. At last, we use a full connection to gather all features and output a predicted RF.  

\subsection{Training Network Parameters}

Before optimizing network parameters, we prepare a large scale of training data. Each training data, that composes of features and a RF label, is generated from a video segment. How to generate features have been mentioned in Chapter II. Meanwhile, the RF label is the best RF parameter, using which the encoder can output compressed video with actual VMAF that equals to a required VMAF value. 

After obtaining a large scale of training data, network parameters can be optimized. The goal of the proposed method is to minimize the error between predicted RF ($R_p$) and ground truth RF label ($R_g$). Therefore, we define loss function as mean square error between $R_p$ and $R_g$: $min\sum|R_p - R_g|^2$. Under this loss function, all network parameters will be optimized by error propagation \cite{rumelhart1985learning}.

Please note, we firstly train a network for the first-pass prediction. When the first network is converged, we use the first network to generate first-pass predicted RF, then encode and measure its corresponding VMAF for each training video segment. Afterwards, we train another network for the second-pass prediction. For the second network, input features include two extra features (a predicted RF and its corresponding VMAF of first-pass prediction and encoding) compared with the first network.

\section{Experiments}

In the the proposed method, RF prediction models are trained for a certain encoder, because RF-VMAF relationships are different for different video encoders. To verify performance of the proposed method, experiments are performed based on x265 encoder \cite{x265}, which is an open-source practical HEVC encoder \cite{sullivan2012overview}. x265 is run with default settings under CRF mode. Meanwhile, the target VMAF value is set as 91 for 1920 $\times$ 1080 videos.

\subsection{Training}

From \href{https://www.bilibili.com/}{bilibili.com}, we collect 500 thousands User Generated Content (UGC) source videos for training and 10 thousands other videos for testing. All these videos have resolutions that are approximate 1920 $\times$ 1080. We separate training video into 2 parts, in where 300 thousands are used to train first-pass RF prediction model and the rest 200 thousands to train second-pass model.

Firstly, we split one-shot segments for each video using scene-cut algorithm of x264 encoder \cite{x264}. For each segment, we generate spatial-temporal and pre-coding features using the method described in \ref{SubSec-Spatial-Tempotal-Features} and \ref{SubSec-Precoding-Features}. Feature extraction processes are accelerated by GPU, and can be run in a speed beyond 6x real-time.

Then, we set target VMAF value as 91 and search ground truth RF label for each training video segment. In detail, we perform many times of encoding and VMAF measuring for each training video segment. In each time, we change RF, encoding video and measure VMAF. Then we find the best RF, which corresponding VMAF equals to the target, as the ground-truth label. Finding RF label consumes about 1 week and large computational resources. 

After obtaining features and RF labels, we can train first-pass RF deep neural network on the first part of training videos. The network is implemented using pyTorch framework, and its parameters are optimized by Adam algorithm with learning rate set as $1e-4$. The training process requires about 12 hours to converge. 

Please note that although searching RF label and training model consumes much time, but once the model is optimized, the inference of the network can be run in only 100 milliseconds, which is ignorable compared with encoding.

We use this model to predict an RF for each segment in the second part of training videos. With the predicted RF, we encode each video segment and measure its actual VMAF. Finally, we can train another RF prediction model based on spatial-temporal, pre-coding and first-pass feedback features. Training method for the second prediction network is the same as the first one. 

\subsection{Testing}

Now, we have two optimized RF prediction models, which are used in first and second pass prediction respectively. For each one-shot segment in a test video, we first predict an RF using the first model. Then we encode the video and measure its actual VMAF. 

If the actual VMAF is within [90, 92], we save the first pass encoded stream and continue to next segment. If the first pass VMAF isn't within [90, 92], we perform a second pass RF prediction and encoding. The second pass actual VMAF measurement is only performed in testing for evaluating model accuracy, but is not performed in practise use, because the second prediction accuracy is pretty high and we can directly save the second pass encoded stream.

\begin{table}[tph!]
\centering
\small
\caption{VMAF accuracy of RF fixed encoding and the proposed method.}
\label{table:vmaf_accuracy}
\begin{tabular}{c|cccc}
\hline
Encoding methods                         & \begin{tabular}[c]{@{}c@{}}Accuracy of\\ $|V_{actual}-91|<1$\end{tabular} & \begin{tabular}[c]{@{}c@{}}Accuracy of\\ $|V_{actual}-91|<2$\end{tabular} & \begin{tabular}[c]{@{}c@{}}Accuracy of\\ $|V_{actual}-91|<3$\end{tabular} & \begin{tabular}[c]{@{}c@{}}Accuracy of\\ $|V_{actual}-91|<4$\end{tabular} \\ \hline
RF fixed encoding             & \textbf{15.90\%}                                                          & 31.68\%                                                                   & 46.08\%                                                                   & 59.70\%                                                                   \\
1st pass encoding of proposed method  & \textbf{45.38\%}                                                          & 74.38\%                                                                   & 88.33\%                                                                   & 95.28\%                                                                   \\
2nd pass encoding of proposed method & \textbf{98.88\%}                                                          & 99.97\%                                                                   & 100\%                                                                     & 100\%                                                                     \\ \hline
\end{tabular}
\end{table}

\subsection{Results}

To evaluate the performance of the proposed method, we calculate the prediction accuracy of first and second pass prediction. Formally, the prediction accuracy is defined as ratio of the number of those testing video segments, which actual VMAF ($V_{actual}$) is within a certain error range around target VMAF (91), to the total number (about 100 thousands) of all testing segments. We analyze 4 error ranges, including $\pm1$, $\pm2$, $\pm3$ and $\pm4$. 

For comparison, we build a comparative encoding method, which encodes all testing segments using a fixed RF. With this fixed RF, the averaged actual VMAF is 91. The results are shown as Table. \ref{table:vmaf_accuracy}.

The experimental results show that RF fixed method only has 15.9\% accuracy, while the proposed method achieves 45.38\% accuracy in first pass prediction and 98.88\% in second pass under $\pm1$ error range. So in the proposed method, only 54.62\% segments require a second pass prediction and encoding. In another word, the proposed method needs only 1 time of VMAF measurement and about 1.55 times of prediction and encoding on average. 

We also draw the VMAF distributions in Fig. \ref{fig:network} for better visualization. Compared with RF fixed method, the proposed method can stably control the actual VMAF of encoding segments close to the target. As as result, the output compressed video of our method has constant VMAF quality between segments.

\begin{figure}[tbp]
\setlength{\abovecaptionskip}{0.cm}
\setlength{\belowcaptionskip}{0.cm}
\centerline{\includegraphics[width=0.8\linewidth]{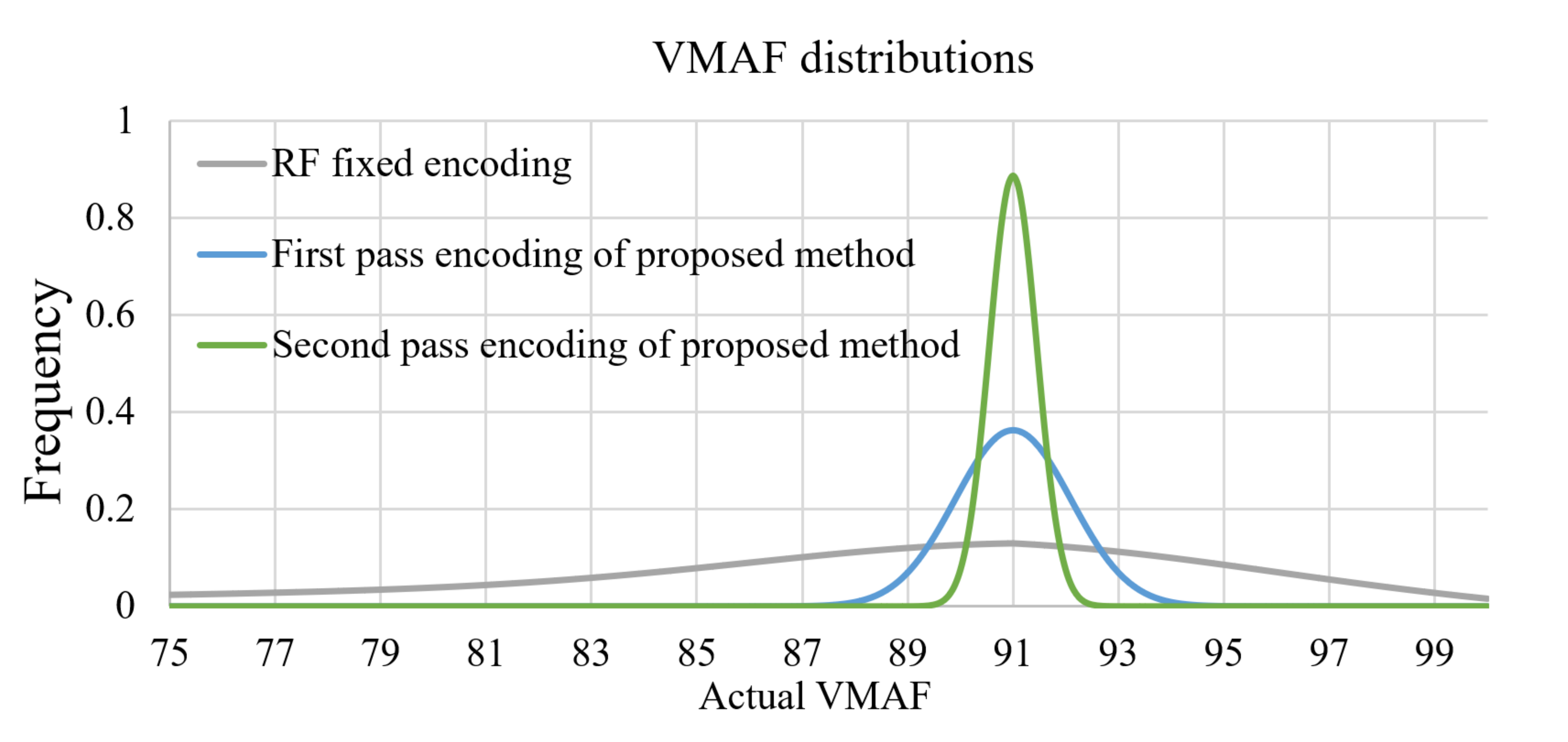}}
\caption{VMAF distributions of RF fixed method and the proposed method. The proposed method can stably control the actual VMAF close to the target}
\label{fig:network}
\end{figure}

\section{Conclusion}

This paper proposed a two-pass rate factor (RF) prediction method based on deep neural networks. In first-pass, an RF is predicted based on spatial-temporal and pre-coding features of video segment. Then video is encoded using the predicted RF and then its VMAF is measured. If first pass VMAF doesn't meet target quality, a second pass prediction is performed using another model, in where results of first pass is added to features. Finally, the proposed method can output compressed video, which actual VMAF can meet target quality at 98.88\% accuracy, using only 1.55 times of encoding on average.

%Bibliography
\bibliographystyle{unsrt}  
\bibliography{predict_crf}

\end{document}